\documentclass{CUP-JNL-DTM}%

\usepackage{graphicx}
\usepackage{multicol,multirow}
\usepackage{amsmath,amssymb,amsfonts}
\usepackage{mathrsfs}
\usepackage{amsthm}
\usepackage{rotating}
\usepackage{appendix}
\usepackage{subcaption}
\usepackage{booktabs}
\usepackage{natbib}
\usepackage{adjustbox}
\usepackage{wrapfig}
\usepackage{ifpdf}
\usepackage[T1]{fontenc}
\usepackage{newtxtext}
\usepackage{newtxmath}
\usepackage{textcomp}
\usepackage{xcolor}
\usepackage{lipsum}
\usepackage{tikz}
\usepackage[colorlinks,allcolors=blue]{hyperref}

\theoremstyle{definition}

\numberwithin{equation}{section}
\newcommand{\cyancirc}{\tikz[baseline=-0.5ex]\draw[cyan,thick] (0,0) circle (1 ex);}
\newcommand{\eg}{\textit{e.g. }}

\jname{Data/Math}
\articletype{ARTICLE TYPE}
\jyear{2024}
\begin{document}

\begin{Frontmatter}

\title[Article Title]{Tree semantic segmentation from aerial image time series}

\author[1,2]{Venkatesh Ramesh}
\author[1,3]{Arthur Ouaknine}
\author[1,3]{David Rolnick}

\authormark{Ramesh \textit{et al}.}

\address[1]{\orgdiv{Mila}, \orgname{Quebec AI Institute}, \orgaddress{\city{Montréal}, \postcode{ H2S 3H1}, \state{Québec},  \country{Canada}} \email{venkatesh.ramesh@mila.quebec}}

\address[2]{\orgdiv{Département d'informatique et de recherche opérationnelle}, \orgname{Université de Montréal}, \orgaddress{\city{Montréal}, \postcode{H3T 1J4}, \state{Québec},  \country{Canada}}}

\address[3]{\orgdiv{School of Computer Science}, \orgname{McGill University}, \orgaddress{\city{Montréal}, \postcode{H3A 2A7}, \state{Québec},  \country{Canada}}.}

% \authormark{Author Name1 et al.}


% \keywords[MSC Codes]{\codes[Primary]{CODE1}; \codes[Secondary]{CODE2, CODE3}}

% \abstract{Forests, crucial to the Earth ecosystems, are disproportionately affected by climate change, influencing their geospatial and species distributions. To effectively monitor forests and mitigate biodiversity loss, it is essential to accurately identify different tree species. In this work, we address this challenge by performing semantic segmentation of trees using an aerial image dataset spanning over a year. We compare models trained on single images versus those trained on time series to assess the impact of tree phenology on segmentation performances. We also introduce a simple convolutional block for extracting spatio-temporal features from image time series, enabling the use of popular pretrained backbones and methods. We leverage the hierarchical structure of tree species taxonomy by incorporating a custom loss function that penalizes predictions at three levels: species, genus, and higher-level taxons. Our findings demonstrate the superiority of our method in exploiting the time series modality and confirm that enriching labels using taxonomic information improves the semantic segmentation performance for forest monitoring.}

\abstract{Earth's forests play an important role in the fight against climate change, and are in turn negatively affected by it. Effective monitoring of different tree species is essential to understanding and improving the health and biodiversity of forests. In this work, we address the challenge of tree species identification by performing semantic segmentation of trees using an aerial image dataset spanning over a year. We compare models trained on single images versus those trained on time series to assess the impact of tree phenology on segmentation performances. We also introduce a simple convolutional block for extracting spatio-temporal features from image time series, enabling the use of popular pretrained backbones and methods. We leverage the hierarchical structure of tree species taxonomy by incorporating a custom loss function that refines predictions at three levels: species, genus, and higher-level taxa. Our findings demonstrate the superiority of our methodology in exploiting the time series modality and confirm that enriching labels using taxonomic information improves the semantic segmentation performance.}

\end{Frontmatter}

\section*{Impact Statement}
This work advances forest monitoring using deep learning on aerial imagery time series. By leveraging phenological information and taxonomic hierarchies, our proposed methods improve tree species segmentation performances. The introduction of a compact spatio-temporal feature extraction module enables the use of pretrained models for this task. Our findings highlight the importance of incorporating temporal data and hierarchical knowledge in forest monitoring and we hope our work will offer valuable insights for biodiversity conservation and climate change mitigation efforts.

\localtableofcontents

\section[Introduction]{Introduction}

Climate change and biodiversity loss in forests are closely intertwined, with each potentially exacerbating the other. As the climate changes, the suitable habitat for many tree species shifts, shrinks, or disappears altogether, leading to changes in forest composition and potential biodiversity loss \citep{Lenoir2008, Allen2010}. Conversely, biodiversity loss in forests can reduce their ability to absorb and store carbon, further contributing to climate change. Different tree species have varying tolerances to changes in temperature, precipitation, and other environmental factors. As a result, climate change can cause variable phenological changes \citep{cite2_1}, shifts in species distribution \citep{cite2_5}, and differential growth responses due to increased atmospheric $\text{CO}_{2}$ \citep{cite2_2, cite2_3}. Phenology in trees refers to the timing of seasonal events such as leaf emergence, color change, and leaf fall. These cyclical changes are influenced by environmental factors like temperature and day length, and often vary between tree species. Understanding phenological patterns can potentially enhance our ability to distinguish between tree species and monitor their responses to environmental changes.

Increasingly, deep learning-based methods, alongside remote sensing applications (\eg land-use and land-cover mapping \citep{cite3_1, cite3_2, cite3_3, forest_health_mapping}, change detection \citep{cite3_5}), have helped with advancing the field of forest monitoring in tree species classification \citep{Fricker2019}, biomass estimation \citep{Zhang2019} and tree crown segmentation \citep{forest_mapping_teja, Weinstein2020}. 

The use of temporal data as inputs of these methods has also shown successes in other tasks such as crop mapping \citep{pixel_set, vit4sits, revisiting_sits} and forest health mapping  \citep{forest_health_mapping}. 
Semantic segmentation of tree crowns is a crucial task in forest monitoring as it provides valuable information about forest composition and health. 
It could be further explored by leveraging time series inputs to learn phenological changes that occur between seasons according to each tree species throughout the years. 

In this work, we evaluate multiple models on the task of tree crown segmentation using a rich dataset recorded in the Laurentides region of Québec, Canada \citep{Cloutier2023}. Among the numerous datasets available for tree crown segmentation \citep{ouaknine2023openforest}, we chose this one for its unique characteristics: high-resolution time series data and a number of closely-related classes. 
This allows us to investigate the impact of phenological (seasonal) changes on tree species identification and assess the ability of the model to distinguish between closely related species. 

To this end, we employ state-of-the-art models in semantic segmentation for single-image and time series segmentation. Additionally, we introduce a lightweight module to extract spatio-temporal features from a time series input, allowing it to be used with backbones that typically operate on single images. 
The dataset we use lacks fine-grained species-level labels for all trees, as it is challenging to accurately identify tree species at a granular level. As a result, it is often easier to identify them on a coarser (genus or family) level. To address this, we propose a custom hierarchical loss function that incorporates labels from all three levels (species, genus, and family) and penalizes incorrect predictions at each level. Overall, the key takeaways from our work can be summarized as follows:
\begin{itemize}
\item We introduce a simple yet effective module for extracting spatio-temporal features, enabling the use of pretrained models for segmenting tree crowns with time series.
\item Our results demonstrate the importance of time series data in identifying tree species, particularly when considering phenological changes.
\item We evaluate models that perform well across taxonomic hierarchies by leveraging a custom hierarchical loss function.
\end{itemize}

\section[Related Work]{Related Work}
\label{sec:related_work}

\subsection{Semantic segmentation}
Deep learning applications for computer vision have been widely explored over the years, including various methods based on convolutional neural networks (CNNs) such as Fully Convolutional Networks (FCNs) \citep{long2015}, U-Net \citep{Unet}, and DeepLab \citep{deeplabv3}.
 
The `dilated' (also named `atrous') convolution \citep{deeplabv3, yu2016} has been introduced to increase the receptive field of CNNs, while attention mechanisms \citep{oktay2018, fu2019} have been incorporated to focus on relevant regions. Multi-scale and pyramid pooling approaches, such as PSPNet \citep{pspnet} and DeepLabV3+ \citep{deeplabv3plus}, have been employed to capture context at different scales. 
Specific methods have also been designed to exploit temporal information for semantic segmentation, \eg with 3D U-Net \citep{Unet3d} and V-Net \citep{milletari2016}.

Recently, transformer-based models have gained popularity in semantic segmentation, showing impressive results, \eg Mask2Former \citep{Mask2Former} combining strengths of CNN-based and transformer-based architectures. It employs a hybrid approach with a CNN backbone for feature extraction and a transformer decoder for capturing global context and generating high-resolution segmentation masks. Other transformer-based models, such as SETR \citep{zheng2021}, TransUNet \citep{chen2021}, and SegFormer \citep{xie2021}, have also been proposed, leveraging the self-attention mechanism to capture long-range dependencies and global context effectively. These latter methods  have demonstrated competitive or improved performances on various semantic segmentation benchmarks compared to traditional CNN-based models.

\subsection{Satellite image time series (SITS)}
Leveraging the temporal information with satellite and aerial imagery provides information on land dynamics and phenology. Researchers have used convolutional neural networks (CNNs) in temporal convolutions for land cover mapping \citep{Lucas2021} and crop classification \citep{Ruwurm2018}. 
Attention-based methods have been used for encoding time series, which have shown to be well-suited for satellite imagery \citep{utae, pixel_set, Ruwurm2023}.
More recently, transformer-based methods have proven their merit using SITS with self-supervised learning exploiting unlabelled data to improve performance on downstream tasks \citep{vit4sits, presto, satmae, scalemae}. 

A recent method has also proposed a new encoding scheme for SITS in order to fit popular pretrained backbones rather than creating task-specific architectures \citep{revisiting_sits}.

\subsection{Forest monitoring}

Deep learning methods have helped advance the field of vegetation monitoring using remote sensing, including both satellite and aerial imagery \citet{Kattenborn2021}, enabling progress in forest monitoring for accurate and efficient analysis at scale \citep{Bae2019, Reichstein2019, Nguyen2024}. 
Such models have achieved state-of-the-art performance in classifying tree species from high-resolution remote sensing imagery \citep{Fricker2019, Onishi2021}. 

Mapping deforestation at large scale using satellite imagery has also been explored \citep{OrtegaAdarme2020, maretto}.
Computer vision and remote sensing have also been leveraged in applications to plant phenology \citep{Katal2022}.
Global vegetation phenology has been modelled with satellite imagery alongside meteorological variables as inputs of a 1D CNN \citep{Zhou2021}. Automated monitoring of forests have also been investigated to accurately identify key phenological events \citep{Cao2021, Song2022, Wang2023}.

Deep learning-based segmentation methods have been applied to automatically delineate individual tree crowns from high-resolution remote sensing imagery \citep{Weinstein2020, Brandt2020, forest_mapping_teja, Li2023}. In a similar vein, a U-Net architecture has been used for fine-grained segmentation of plant species using aerial imagery \citep{Kattenborn2019}. 
A foundation model trained on datasets from multiple sources is also able to perform decently on a variety of downstream tasks for forest monitoring, including classification, detection and semantic segmentation \citep{fomo_bench}.

\subsection{Hierarchical losses}

Hierarchical loss functions have been extensively explored in various tasks to leverage the inherently hierarchical structure of object classes. By incorporating information from different levels of granularity, such loss functions aim to improve the ability of the model to make fine-grained distinctions and enhance overall performance.
For classification tasks, a curriculum-based hierarchical loss gradually increasing the specificity of the target class was explored by \citet{hierarchical_curriculum}.
Similarly, a loss function evaluated at multiple operating points within the class hierarchy has helped to capture information at various levels of this hierarchy \citep{hierarchical_operating_points}.
In contrast, one may encourage the model to make better mistakes by assigning different weights to the misclassified samples based on their position in the hierarchy, promoting more semantically meaningful errors \citep{hierarchical_making_better_mistakes}.

Hierarchical loss functions have also been applied to object detection \citep{hloss_od1, hloss_od2} and semantic segmentation \citep{hloss_semseg_1, hloss_semseg_2, hloss_semseg_3} demonstrating the effectiveness of incorporating a more structured and informative signal during the learning process.

\section{Dataset} \label{Dataset}

\begin{figure}[tbp]
\centering
\begin{subfigure}{0.45\textwidth}
\centering
\includegraphics[width=\textwidth]{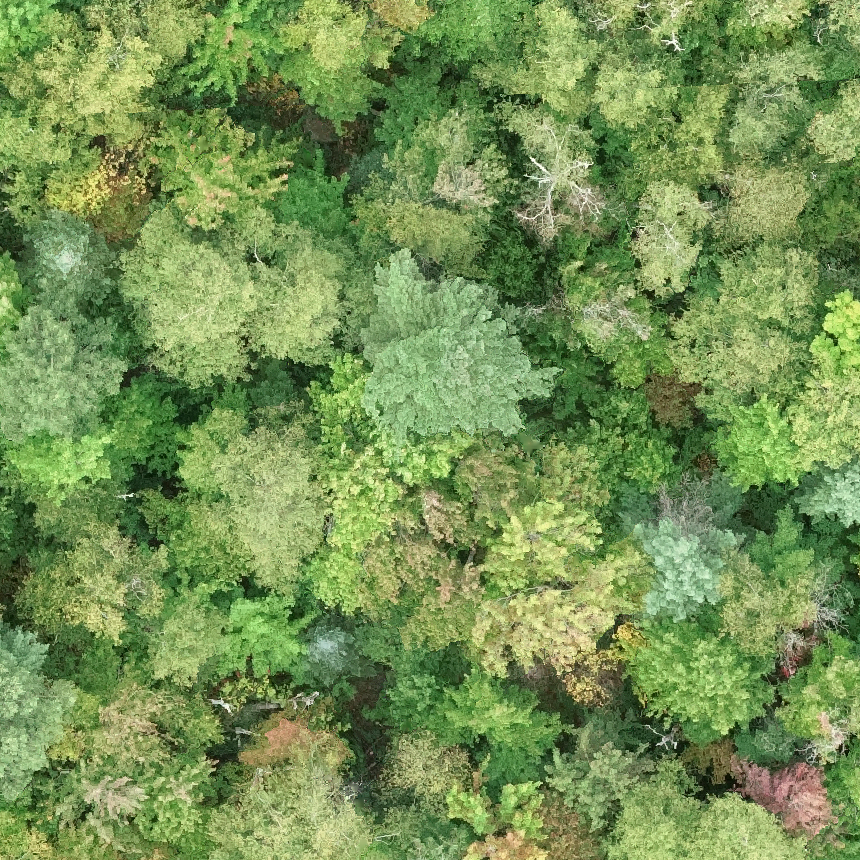}
\caption{Sample from 2nd September 2021.}
\label{fig:image1_sample}
\end{subfigure}
\hfill
\begin{subfigure}{0.45\textwidth}
\centering
\includegraphics[width=\textwidth]{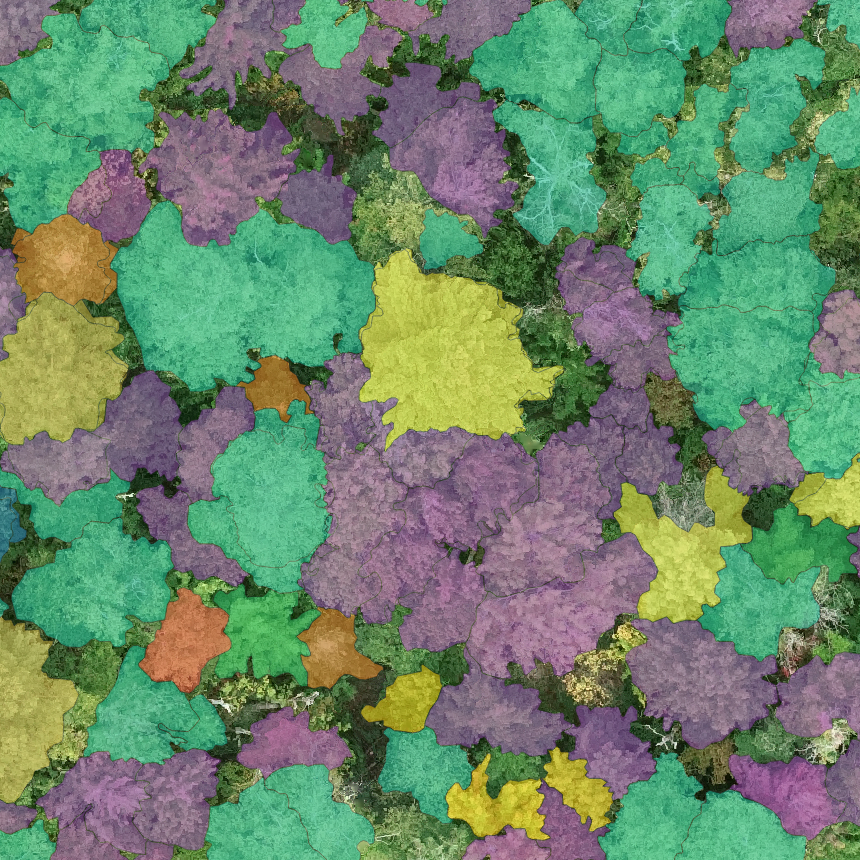}
\caption{Labels overlaying the sample image.}
\label{fig:image1_label}
\end{subfigure}
\caption{\textbf{Example of an annotated sample from the studied dataset}. The image \ref{fig:image1_sample} shows a scene captured on September 2nd, while the image \ref{fig:image1_sample} overlays the tree species labels on the same scene. Each tree species is represented by a distinct color as seen in Table~\ref{tab:tree-abbreviations}.}
\label{fig:sample_images}
\end{figure}

The dataset used in our work \citep{Cloutier2023} consists of high-resolution RGB imagery from unmanned aerial vehicles (UAVs) at seven different acquisition dates over a temperate-mixed forest in the Laurentides region of Québec, Canada during the year 2021. The acquisitions were conducted monthly from May to August, with three additional acquisitions in September and October to capture colour changes during autumn. The dataset contains a total of 23,000 individual tree crowns that were segmented and annotated, mostly at the species level, with 1,956 trees annotated only at the genus level due to the difficulty in accurately identifying species-level labels. This dataset offers a unique combination of time series data and a large number of fine-grained tree species. This allows us to leverage the temporal information  to investigate the impact of phenological changes on tree species identification. An example of this dataset is shown in Figure~\ref{fig:sample_images}.

\begin{figure}[tbp]
\centering
\begin{subfigure}{0.45\textwidth}
\centering
\includegraphics[width=\textwidth]{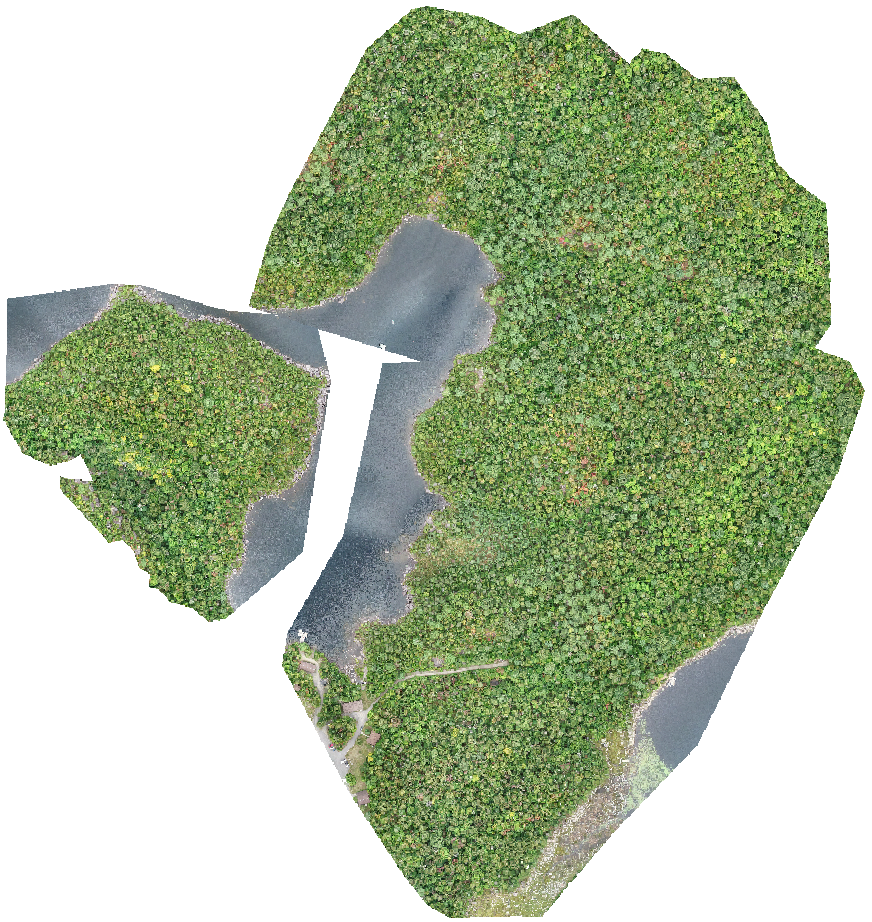}
% \caption{An example image from the dataset taken on 2nd September.}
% \label{fig:image1}
\end{subfigure}
\hfill
\begin{subfigure}{0.45\textwidth}
\centering
\includegraphics[width=\textwidth]{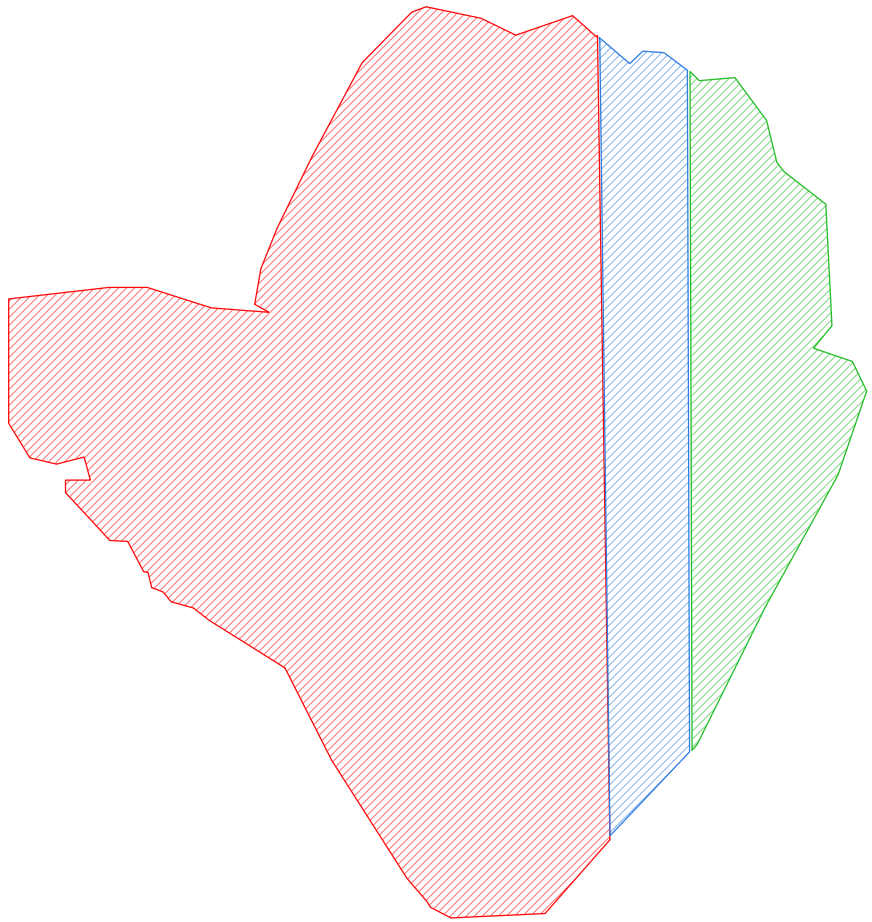}

\end{subfigure}
\caption{\textbf{Spatial splits of the dataset.} The image on the left depicts the entire region where the aerial imagery was captured, while the image on the right shows the different subregions used  to train, evaluate and test models.The training region is represented by {\color{red}\rule{0.3cm}{0.3cm}}, the validation region in {\color{blue}\rule{0.3cm}{0.3cm}}, and the test region in {\color{green}\rule{0.3cm}{0.3cm}}. To prevent data leakage between the subsets, a buffer tile is omitted between the adjacent regions. This spatial partitioning ensures that the model's performance is assessed on geographically distinct areas, simulating real-world scenarios where the model would be applied to unseen locations.}
\label{fig:train_val_test}
\end{figure}

We create splits which are separated spatially for training, validation, and testing while ensuring an equal distribution of the selected classes in each split. The spatial splits are used to evaluate the performances of each model on geographically distinct areas and to simulate real-world scenarios, \eg applications to unseen locations. The splits that we used are illustrated in Figure~\ref{fig:train_val_test}. A tile is skipped between the three sets to prevent data leakage, ensuring the model avoids spatial autocorrelation between adjacent areas. For our experiments, we use an image size of $768 \times 768 \times 3$, providing sufficient spatial context to include multiple tree crowns and to learn relationships between different regions in the image. The labels are annotated using recordings from September 2 as reference (representing a date before most leaves change colour), which is also used as the input for our single-image models. For the models that take time series as input, we select one image from June, two from September, and one from October to reduce redundant information, as most phenological changes occur between September and October.

\begin{figure}[th]
\centering
\begin{minipage}{0.48\textwidth}
\centering
\includegraphics[width=\textwidth]{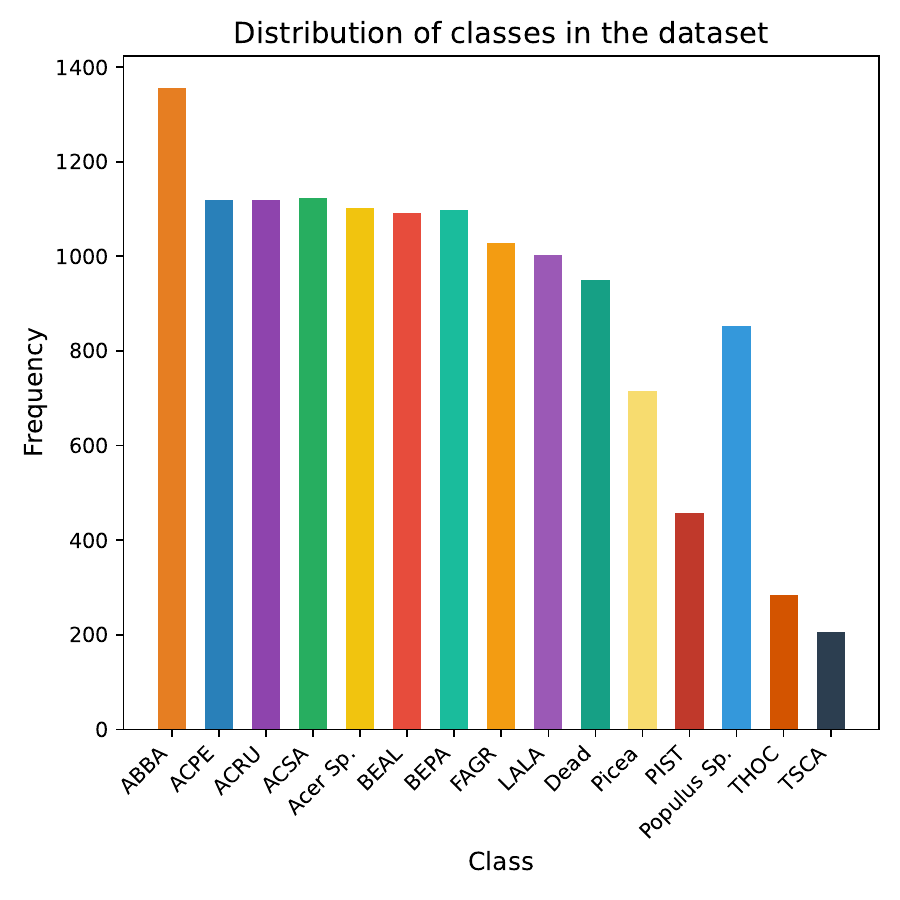}
\captionof{figure}{\textbf{Distribution of the selected classes in the dataset}. We observe that there is a substantial difference in the frequency of occurrence of each tree species. The common and scientific names used for the abbreviations are detailed in Table~\ref{tab:tree-abbreviations}.}
\label{fig:distribution}
\end{minipage}
\hfill
\begin{minipage}{0.48\textwidth}
\centering
\resizebox{1.\textwidth}{!}{
\begin{tabular}{ll}
\toprule
\textbf{Common Name (Scientific Name)} & \textbf{Abbreviation} \\
\midrule
Balsam fir (Abies balsamea) & \colorbox[HTML]{E67E22}{\textcolor{white}{ABBA}} \\
Striped maple (Acer pensylvanicum) & \colorbox[HTML]{2980B9}{\textcolor{white}{ACPE}} \\
Red maple (Acer rubrum) & \colorbox[HTML]{8E44AD}{\textcolor{white}{ACRU}} \\
Sugar maple (Acer saccharum) & \colorbox[HTML]{27AE60}{\textcolor{white}{ACSA}} \\
Maple (Acer sp.)  & \colorbox[HTML]{F1C40F}{\textcolor{black}{Acer}} \\
Swamp birch (Betula alleghaniensis) & \colorbox[HTML]{E74C3C}{\textcolor{white}{BEAL}} \\
Paper birch (Betula papyrifera) & \colorbox[HTML]{1ABC9C}{\textcolor{white}{BEPA}} \\
American beech (Fagus grandifolia) & \colorbox[HTML]{F39C12}{\textcolor{white}{FAGR}} \\
Tamarack (Larix laricina) & \colorbox[HTML]{9B59B6}{\textcolor{white}{LALA}} \\
Dead tree & \colorbox[HTML]{16A085}{\textcolor{white}{DEAD}} \\
Spruce (Picea sp.)  & \colorbox[HTML]{F7DC6F}{\textcolor{black}{Picea}} \\
Eastern white pine (Pinus strobus) & \colorbox[HTML]{C0392B}{\textcolor{white}{PIST}} \\
Aspen (Populus sp.) & \colorbox[HTML]{3498DB}{\textcolor{white}{Populus}} \\
Northern white-cedar (Thuja occidentalis) & \colorbox[HTML]{D35400}{\textcolor{white}{THOC}} \\
Eastern hemlock (Tsuga canadensis) & \colorbox[HTML]{2C3E50}{\textcolor{white}{TSCA}} \\
\bottomrule
\end{tabular}
}
\vspace{7pt}
\captionof{table}{\textbf{Tree species names and their abbreviations}. The color we use to depict each species is highlighted in the second column and is consistent for all the plots and figures.}
\label{tab:tree-abbreviations}
\end{minipage}
\end{figure}

As a design choice, we ignore classes with less than 50 occurrences throughout the dataset, leaving us with a total of 15 classes, excluding the background class. This ensures the selected classes have sufficient samples in each split in order to effectively train and evaluate each model. The tree species distribution is illustrated in Figure~\ref{fig:distribution}. 

The dataset is split into train, validation and testing sets with 63\%, 16\%, and 21\% of the samples, respectively.   
Given that this dataset has a mix of coarse (genus) and fine-grained (species) labels, we leverage this information to create a complete taxonomy of the classes used, as seen in Figure~\ref{fig:taxonomy}. 

This taxonomic hierarchy is incorporated in our proposed loss function as detailed in Sec.~\ref{HLoss}.

\section{Methods}
\label{sec:method}
In this section, we provide more details on the methods used to perform semantic segmentation either with single image or time series inputs. We will also describe the proposed hierarichal loss used to exploit the tree label taxonomy.

\subsection{Single image semantic segmentation}
The single image semantic segmentation experiments are conducted with diverse methods detailed in the following sections.

\subsubsection{U-Net}
U-Net \citep{Unet} is a widely adopted convolutional neural network (CNN) architecture \citep{Dong2017, Falk2018, Li2018} designed for efficient image segmentation tasks. The architecture consists of an encoder path and a decoder path, which together form a U-shaped structure. The encoder path follows the typical structure of a CNN, consisting of successive CNN layers, rectified linear units (ReLU), and max-pooling operations, which gradually reduce the spatial dimensions while increasing the number of feature maps. The decoder path utilizes transposed convolutions to upsample the feature channels, enabling the network to construct segmentation maps at the original input resolution. The U-Net architecture uses skip connections \citep{resnet} to concatenate feature maps from the encoder path with the corresponding upsampled feature maps in the decoder path.

\subsubsection{DeepLabv3+}
The DeepLabv3+ architecture \citep{deeplabv3plus} is an image segmentation method built upon strengths of pyramid pooling with an encoder-decoder structure \citep{deeplabv3}. The encoder module of the DeepLabv3+ utilizes `dilated' (also named `atrous') convolutions to extract dense feature maps at multiple scales with larger receptive fields while keeping the computation costs lower. 
The encoder incorporates atrous spatial pyramid pooling (ASPP), which applies atrous convolutions with different dilation rates in parallel to further capture multi-scale context \citep{deeplabv3}. 

The decoder module of the DeepLabv3+ combines the output of the encoder with low-level features from the encoder. This information is refined with $3 \times 3$ convolutions to produce the final output segmentation maps.

\subsubsection{Mask2Former}
The Mask2Former architecture \citep{Mask2Former} is a versatile method that applies binary masks to focus attention only on the areas with foreground features. 
The architecture consists of three parts: a backbone network, a pixel decoder, and a transformer decoder. Universal backbones (ResNet \citep{resnet} or Swin Transformer \citep{SwinTransformer}) are used to extract features from the input image. The low-resolution features are then used in a pixel decoder and upsampled to higher resolution. The masked attention is finally applied on the pixel embeddings in the transformer decoder.

To reduce the computational burden of using high-resolution masks, the transformer decoder processes the multi-scale features per resolution one at a time. The Mask2Former architecture performs well across a variety of tasks like semantic, instance, and panoptic segmentation, which makes it a popular choice.

\subsection{Time series semantic segmentation}
We compare various methods for semantic segmentation with time series data, including 3D-UNet \citep{Unet3d}, specialized for medical images, and U-TAE \citep{utae} specialized for SITS. 
Additionally, we propose a simple, yet effective, module composed of 3D convolutional layers, referred to as `Processor', to preliminarily process the time series and use its representation as input for mainstream single-image segmentation methods.

\subsubsection{3D-UNet}

The 3D-UNet method \citep{Unet3d} is composed of successive 3D convolutions with a $3 \times 3 \times 3$ kernel, followed by batch normalization and a leaky ReLU activation. The 3D-UNet downsampling part is composed of five blocks, separated by spatial downsampling after the second and fourth blocks. The upsampling part consists of 5 blocks with transposed convolutions while features from the downsampling part are concatenated similarly than U-Net \citep{Unet}.

\subsubsection{U-TAE}

The U-TAE architecture \citep{utae} has been introduced for panoptic segmentation of SITS. It consists of three main parts: a multi-scale spatial encoder, a temporal encoder, and a convolutional decoder that produces a single feature map with the same spatial resolution as the input. The sequence of images is processed in parallel by the spatial encoder, and the temporal attention encoder (TAE) is applied at the lowest resolution features to generate attention masks. These masks are interpolated and applied to each feature map, allowing the extraction of spatial and temporal information at multiple scales. The decoder uses a series of transposed convolutions, ReLU, and batch normalization layers to produce the final feature map.

\subsubsection{Processor module}
Our proposed Processor module is composed of 3D convolutions and designed to extract spatio-temporal features from time series data, enabling the use of pretrained models for semantic segmentation.  % The motivation behind the Processor architecture is to leverage the power of convolutions in capturing both spatial and temporal information while maintaining the flexibility to use established pretrained models. 
The motivation behind the Processor architecture is to capture spatio-temporal patterns while maintaining the spatial resolution to fit established models pretrained on single-image datasets.
This approach differs from task-specific models relying on specialized architectures for processing time series data in particular contexts, such as land use and land cover mapping \citep{vit4sits, utae}.

The module is composed of two 3D convolutional layers. The first layer has a kernel size of $3 \times 3 \times 3$, followed by a second layer with a kernel size of $2 \times 3 \times 3$. The padding in these layers is set to $(0, 1, 1)$, and the number of output channels is set to 32 and 64 respectively. 
This configuration will collapse the temporal dimension of the input while simultaneously increasing the number of channels. 

Since the kernel sizes are designed for a specific time series length, they must be adjusted for a different application, yet our lightweight module is easily trainable from scratch.

Formally, let $\mathbf{x} \in \mathbb{R}^{T \times C \times H \times W}$ be an input time series, where $T$ is the length of the time series, $C$ the number of channels of each image, $H$ and $W$ their respective height and width dimensions. Our Processor module $p_\Theta(.)$, parameterized by $\Theta$, can be used prior to any semantic segmentation model $f_\theta$  parameterized by $\theta$, via $f_\theta(p_\Theta(\mathbf{x}))$. 
To evaluate the effectiveness of our approach, we used the Processor alongside U-Net and DeepLabv3+. The results of our experiments are detailed in Section \ref{sec:results}. 

\begin{figure}[tbp]
    \centering
    \includegraphics[height=4.5cm, width=\textwidth]{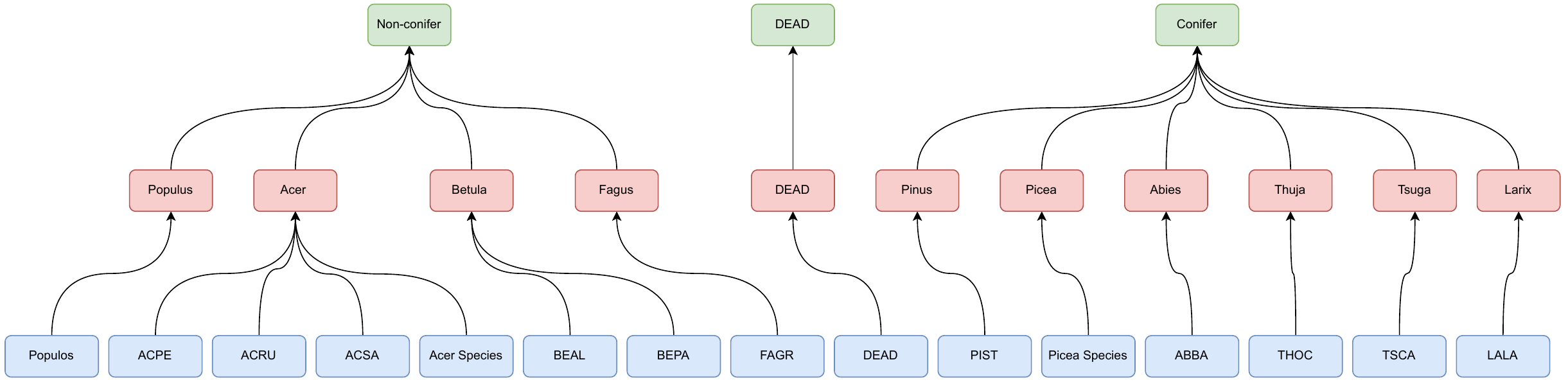}
    \caption{\textbf{Taxonomic hierarchy of tree species.} 
    %The figure shows the taxonomic hierarchy of the tree species used in this study organized into three levels: species (blue), genus (red) and higher-level taxon (green). 
    The hierarchical structure is visually represented using a tree diagram. 
    Blue nodes represent the species level, the most fine-grained classification in the hierarchy. Red nodes denote the genus level, which groups together closely related species. Finally, green nodes group the higher-level taxon, the broadest classification level, which encompasses multiple genera and families. This structure of labels allows the models to learn more comprehensive relationships between different tree species at multiple levels of granularity. The full names of each abbreviation are detailed in Table~\ref{tab:tree-abbreviations}.}
    \label{fig:taxonomy}
\end{figure}

\subsection{Hierarchical loss}\label{HLoss}
\begin{figure}[t!]
\centering
\begin{subfigure}{0.22\textwidth}
\centering
\includegraphics[width=\linewidth]{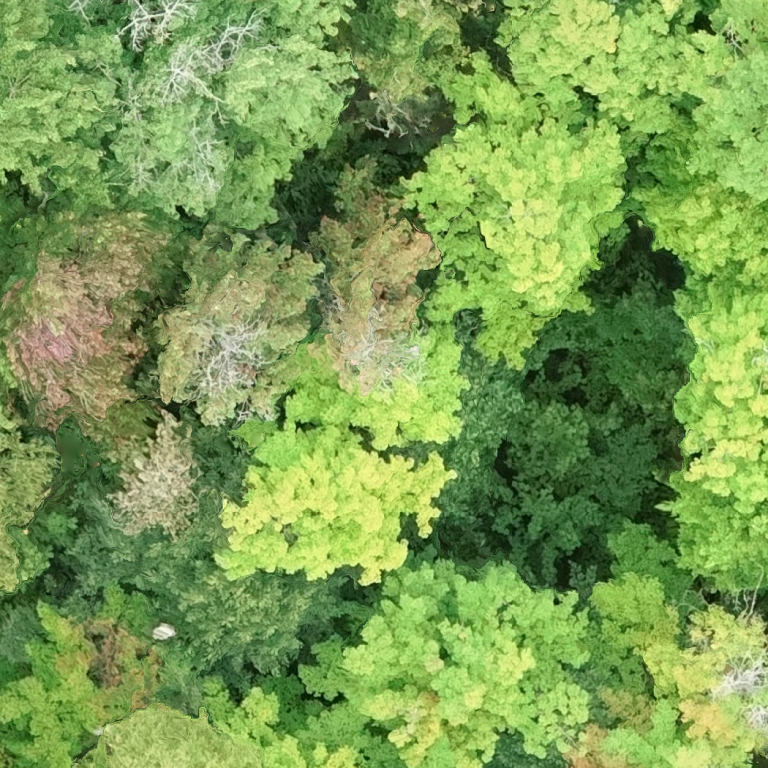}
\caption{Image}
\label{input_image}
\end{subfigure}\hfill
\begin{subfigure}{0.22\textwidth}
\centering
\includegraphics[width=\linewidth]{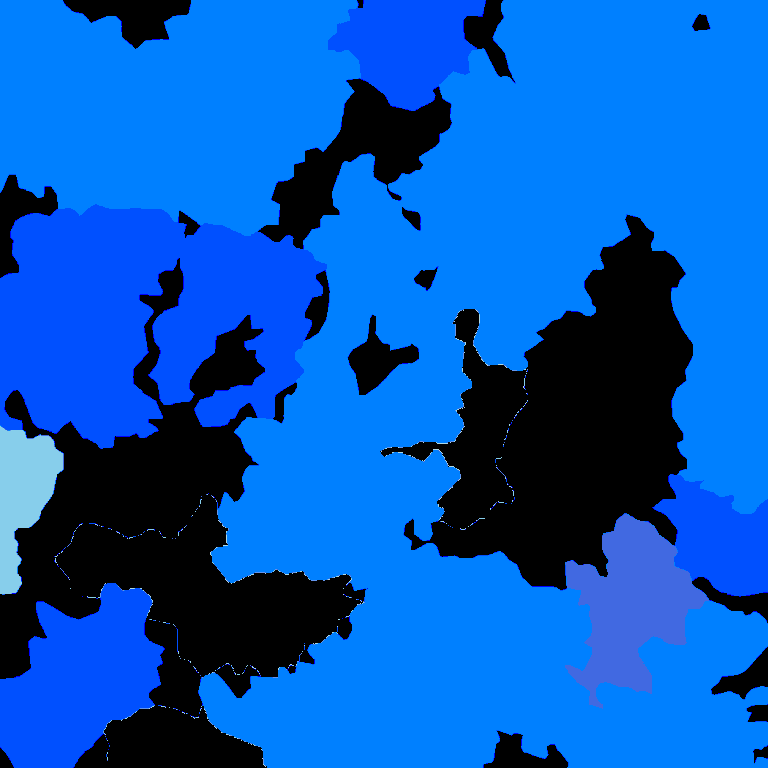}
\caption{Species}
\label{species_channel}
\end{subfigure}\hfill
\begin{subfigure}{0.22\textwidth}
\centering
\includegraphics[width=\linewidth]{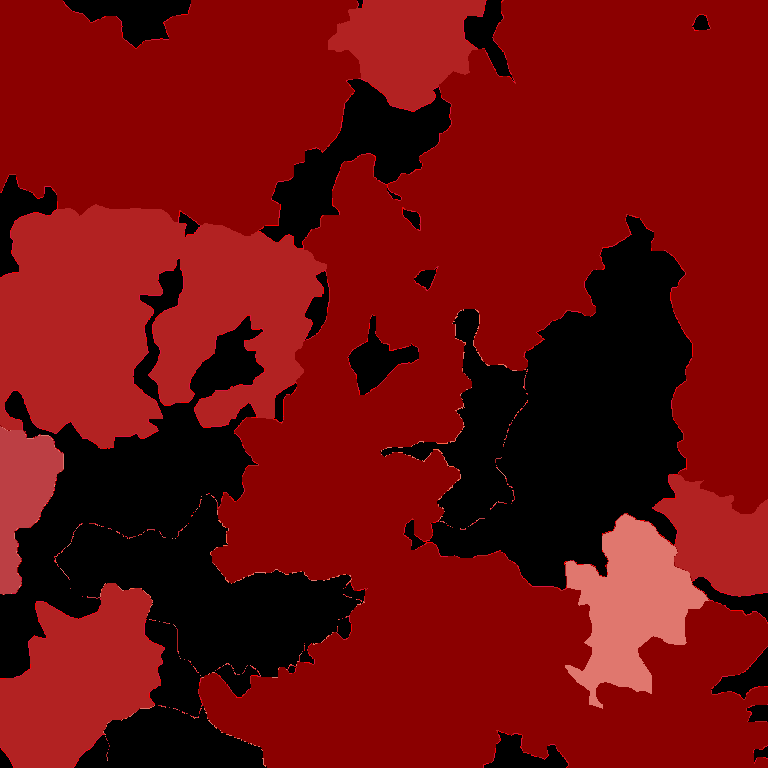}
\caption{Genus}
\label{genera_channel}
\end{subfigure}\hfill
\begin{subfigure}{0.22\textwidth}
\centering
\includegraphics[width=\linewidth]{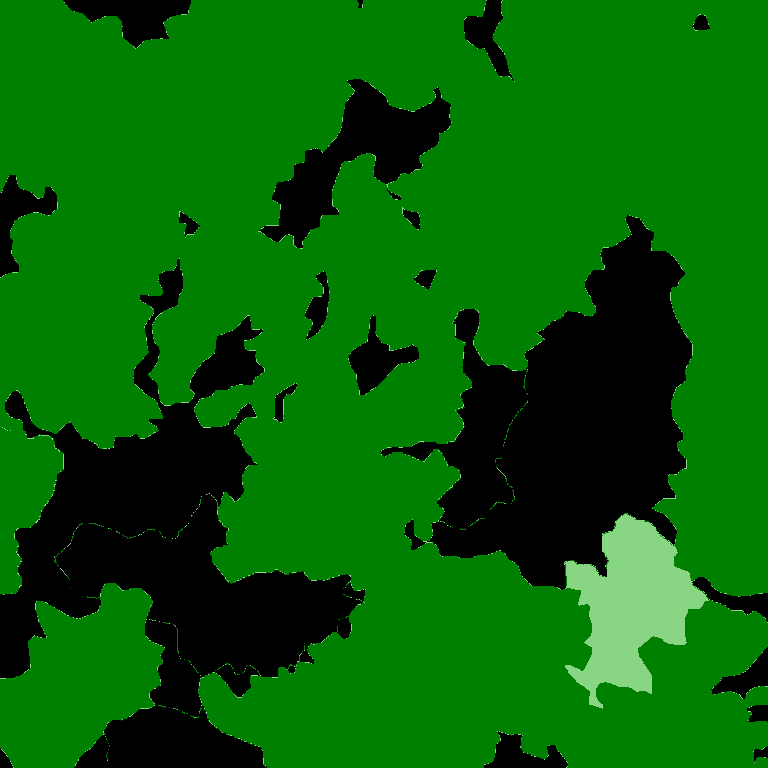}
\caption{Higher-level taxon}
\label{hl_channel}
\end{subfigure}
\caption{\textbf{Example of the proposed three-level hierarchical label structure.}
The labels are concatenated to form semantic segmentation masks where each channel correspond to a specific taxonomic level: species \ref{species_channel}, genus \ref{genera_channel} and higher-level taxon \ref{hl_channel}. In this example, there are three classes at the species and genus level. However, the higher-level taxon only has two classes due to the aggregation of different trees under one class. Note that the colors used in this image do not conform to the color code shown in Table~\ref{tab:tree-abbreviations}.}
\label{fig:three_level_example}
\end{figure}

This section details the proposed hierarchical loss that leverages information about taxonomic hierarchies of tree species, genus and families.
%As mentioned in Section \ref{Dataset}, the dataset we use has a mix of finer (species) and coarser (genus) level labels. 
The dataset detailed in Section~\ref{Dataset} groups a mix of finer (species) and coarser (genus) level labels. 
% The taxonomic structure of these labels offers an avenue to leverage that information to train a model that performs well across the taxonomic hierarchy. 
The taxonomic structure of these labels offers an opportunity to train a model while benefiting from such hierarchical structure.
% To use the hierarchical loss, we first need to generate labels for three levels: species, genus, and higher-level taxon. 
To exploit this hierarchy, we extend each label to multiple levels: species, genus, and higher-level taxon.
The taxonomic hierarchy is illustrated in Figure~\ref{fig:taxonomy}, and 
a visual example of these labels is illustrated in Figure~\ref{fig:three_level_example}. 

During training, the model predicts only the species level labels for each pixel.
These softmax probabilities at species level are then aggregated according to our knowledge of the label taxonomies (see Figure~\ref{fig:taxonomy}) to generate first the genus level predictions (see Equation~\ref{eq:genus_level}) and second the higher-level predictions (see Equation~\ref{eq:family_level}). 

Note that our implementation of the hierarchical loss differs from certain related work presented in Section~\ref{sec:related_work}, where classes at all levels are predicted separately to compute the loss \citep{Turkoglu2021}.

Formally, let $\textbf{x} \in \mathbb{R}^{C \times H \times W}$ be a training example, $\textbf{y}_S \in \{ 0, 1 \}^{S \times H \times W}$ its one-hot ground truth where $S$ is the number of classes at the species level, and $f_\theta(\textbf{x}) = \textbf{p}_S$ the associated predictions. The cross-entropy loss function at the species level is defined as normal via:
\begin{equation}
\mathcal{L}_\text{species} := - \frac{1}{S} \sum_{s=1}^{S} \sum_{(h, w) \in \Omega} \textbf{y}_S [ h, w, s ] \text{log} \ \textbf{p}_S [h, w, s],
\label{eq:species_level}
\end{equation}
where $\Omega = \llbracket 1, H \rrbracket \times\llbracket 1, W \rrbracket $.
The cross-entropy loss function at the genus level is then computed using the ground truth and predictions at the species level, as:
\begin{align}
\mathcal{L}_\text{genus} & := - \frac{1}{G} \sum_{g=1}^{G} \sum_{(h, w) \in \Omega}  \textbf{y}_G [ h, w, g ] \text{log} \ \textbf{p}_G [h, w, g] \\
& =  - \frac{1}{G} \sum_{g=1}^{G} \sum_{(h, w) \in \Omega} \Big [ \sum_{s=1}^{S_g} \textbf{y}_S [ h, w, s ] \Big ]  \text{log} \Big [ \sum_{s=1}^{S_g} \textbf{p}_S [ h, w, s ] \Big ],
\label{eq:genus_level}
\end{align}
where $G$ is the number of classes at the genus level and $S_g$ is the number of classes at the species level corresponding to a given genus class $g$.
In the same vein, the cross-entropy loss function at the higher-level taxon is also obtained via the ground truth and predictions at the species level, as:
\begin{align}
\mathcal{L}_\text{taxon} & := - \frac{1}{T} \sum_{t=1}^{T} \sum_{(h, w) \in \Omega}  \textbf{y}_T [ h, w, t ] \text{log} \ \textbf{p}_T [h, w, t] \\
& = - \frac{1}{T} \sum_{t=1}^{T} \sum_{(h, w) \in \Omega}  \Big [ \sum_{g=1}^{G_t} \sum_{s=1}^{S_g}  \textbf{y}_S [ h, w, s ] \Big ]  \text{log} \Big [ \sum_{g=1}^{G_t} \sum_{s=1}^{S_g}   \textbf{p}_S [ h, w, s ] \Big ],
\label{eq:family_level}
\end{align}
where $T$ is the number of classes at the higher-level taxon and $G_t$ the number of classes at the genus level corresponding to a given higher-level class $t$.

The hierarchical loss function is formulated as:
\begin{equation}
\mathcal{L_{\text{HLoss}}} = \lambda_1 \cdot \mathcal{L}_\text{species} + \lambda_2 \cdot \mathcal{L}_\text{genus} + \lambda_3 \cdot \mathcal{L}_\text{taxon}, 
\end{equation}
where $\lambda_1$, $\lambda_2$, and $\lambda_3$ are the weights for the species, genus, and higher-level taxon losses respectively, and $\mathcal{L}_\text{species}$, $\mathcal{L}_\text{genus}$, and $\mathcal{L}_\text{taxon}$ are the corresponding cross-entropy losses. 

We set empirically $\lambda_1 = 1$, $\lambda_2 = 0.3$, and $\lambda_3 = 0.1$ since we observed that giving more weight to the species-level loss helps the model to prioritize the fine-grained predictions while still benefiting from the hierarchical information. However, we have not attempted to fully optimize these values.

\section{Experiments}

\subsection{Experimental setup}

All methods detailed in Section~\ref{sec:method} have been trained with normalized input data, either with the means and standard deviations of our dataset to train models from scratch, or with statistics of the datasets used for pretraining for models based on MS-COCO and ImageNet weights.

We employ the Adam optimizer \citep{adam_optimizer} for all models except Mask2Former, which is trained with the AdamW optimizer \citep{adamw_optimizer} to maintain consistency with the original training methodology. 
We trained all models with an learning rate of $1\text{e}-4$ with exponential learning rate decay for 300 epochs. 

We included rotation (in multiples of 90\textdegree) with horizontal flips as data augmentation to enhance the diversity of the training data. 

The batch sizes used for each model are detailed in Table~\ref{tab:model-batch-sizes}. These were set to the largest size that could fit within a NVIDIA RTX 8000 GPU. 

We train our models either using our proposed hierarchical loss, noted HLoss, and described in Section \ref{HLoss}, or using a combination of dice and cross-entropy losses, noted Dice$+$CE.
The latter is a popular choice for segmentation tasks since the dice loss measures the overlap between the predicted and ground truth masks, while the cross-entropy loss quantifies the dissimilarity between the predicted and true class probabilities. We trained the Mask2Former model with the loss function proposed by its authors \citep{Mask2Former}. This loss function improves the training efficiency by randomly sampling a fixed number of points in the labels and predictions.
\begin{wrapfigure}{r}{0.5\textwidth}
\centering
\begin{tabular}{lc}
\toprule
\multicolumn{1}{c}{\textbf{Model}} & \textbf{Batch Size} \\
\midrule
U-TAE & 4 \\
Unet-3D & 6 \\
Processor+U-Net & 16 \\
Processor+DeepLabv3+ & 16 \\
U-Net & 16 \\
DeepLabv3+ & 16 \\
Mask2former & 16 \\
\bottomrule
\end{tabular}
\caption{\textbf{Batch sizes used for training}.}
\label{tab:model-batch-sizes}
\end{wrapfigure}

The loss weighing scheme and other implementation details are kept consistent with original implementation to ensure a fair comparison. Note that we did not run Mask2Former with HLoss and Dice$+$CE loss as the training would be much more computationally expensive, resulting in a smaller batch size.

The performances of our models are evaluated with the Intersection over Union (IoU) metric, also known as the Jaccard index, measuring the overlap between the predicted and ground truth masks. Letting $A$ and $B$ be two sets, the IoU score is defined as:

\begin{equation}
\text{IoU}(A, B) := \frac{|A \cap B|}{|A \cup B|} = \frac{|A \cap B|}{|A| + |B| - |A \cap B|}.
\end{equation}

The mean IoU (mIoU) is computed by averaging the IoU scores across all classes. This metric provides a comprehensive assessment of the segmentation performances of a model, taking into account both the precision and recall.

\subsection{Experiment Configuration}

% To evaluate the performances of our models in a thorough manner, we conduct a comprehensive set of experiments:
We conduct a comprehensive set of experiments to thoroughly evaluate the performances of the considered methods:
\begin{itemize}
    \item We compared models using either single-image or time series inputs to evaluate the contribution of the phenological information on the tree species segmentation task. The time series are composed of images at four different periods of the year (see Section~\ref{Dataset}). Note that both methods predict segmentation masks corresponding to a single image.
    \item We compared models with two different loss functions to demonstrate the effectiveness of leveraging taxonomic information through the HLoss against a standard combination of loss functions (Dice$+$CE).
    \item We conduct ablation studies to investigate the impact of different pretrained backbones on the segmentation performances. For the CNN-based models, we experiment with ResNet-34, ResNet-50, and ResNet-101 backbones, whereas for the Mask2Former model, we use Swin-T and Swin-S backbones \citep{SwinTransformer}.
\end{itemize}

The results of these experiments are discussed in Section~\ref{sec:results} where we compare results both quantitatively and qualitatively.

\section{Results}
\label{sec:results}

\begin{table}[tbp]
\centering
\begin{tabular}{lcccc}
\toprule
\multicolumn{1}{c}{\textbf{Model}} & \multicolumn{1}{c}{\textbf{Backbone}} & \multicolumn{1}{c}{\textbf{Dice+CE}} & \multicolumn{1}{c}{\textbf{HLoss}} & \multicolumn{1}{c}{\textbf{Mask2Former Loss}} \\
\midrule
\multirow{4}{*}{DeepLabv3+} & ResNet34 & 52.30 $\pm$ 0.40 & \textbf{53.36 $\pm$ 0.09} & --- \\
& ResNet50 & 53.14 $\pm$ 0.08 & \textbf{53.63 $\pm$ 0.20} & --- \\
& ResNet101 & 53.87 $\pm$ 0.45 & \textbf{54.16 $\pm$ 0.38} & --- \\
& ResNet50$^{\dagger}$ & 43.19 $\pm$ 0.45 & \textbf{43.20 $\pm$ 0.06} & --- \\
\midrule
\multirow{4}{*}{U-Net} & ResNet34 & 53.00 $\pm$ 0.10 & \textbf{53.10 $\pm$ 0.16} & --- \\
& ResNet50 & 53.30 $\pm$ 0.16 & \textbf{53.53 $\pm$ 0.46} & --- \\
& ResNet101 & 53.90 $\pm$ 0.18 & \textbf{\textcolor{red}{54.31 $\pm$ 0.48}} & --- \\
& ResNet50$^{\dagger}$ & \textbf{42.66 $\pm$ 0.38} & 42.43  $\pm$ 0.96 & --- \\
\midrule
\multirow{2}{*}{Mask2Former} & Swin-t$^{\dagger\dagger}$ & --- & --- & 47.41 $\pm$ 0.50 \\
& Swin-s$^{\dagger\dagger}$ & --- & --- & 46.61 $\pm$ 0.10 \\
\bottomrule
\end{tabular}
\caption{\textbf{Comparison of single image methods with different losses and backbones.} Performances are compared with IoU averaged over all the classes of the dataset (mIoU) for single image models. The $^{\dagger}$ indicates models trained from scratch without using ImageNet weights \citep{imagenet}. The $^{\dagger\dagger}$ indicates Swin-based models using weights from MS-COCO dataset \citep{ms_coco}. All the results are averaged over three seeds and the best results for a particular backbone is shown in bold text. The best model overall is highlighted in red.}
\label{tab:si_results}
\end{table}

\begin{figure}[b]
    \centering
    \begin{subfigure}{0.22\textwidth}
        \includegraphics[width=\textwidth]{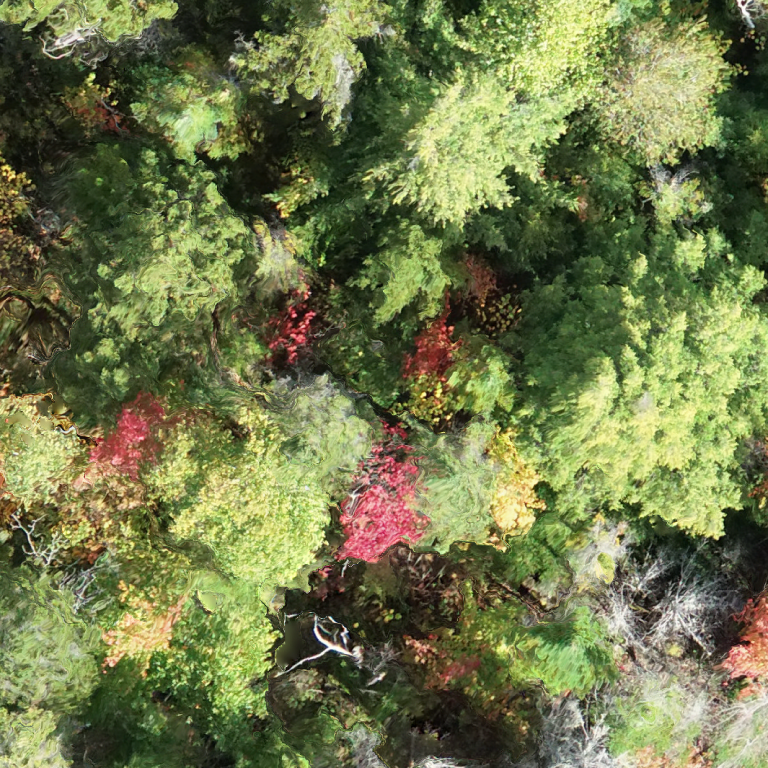}
        \caption{A sample image.}
        \label{fig:results_loss_1}
    \end{subfigure}
    \hfill
    \begin{subfigure}{0.22\textwidth}
        \includegraphics[width=\textwidth]{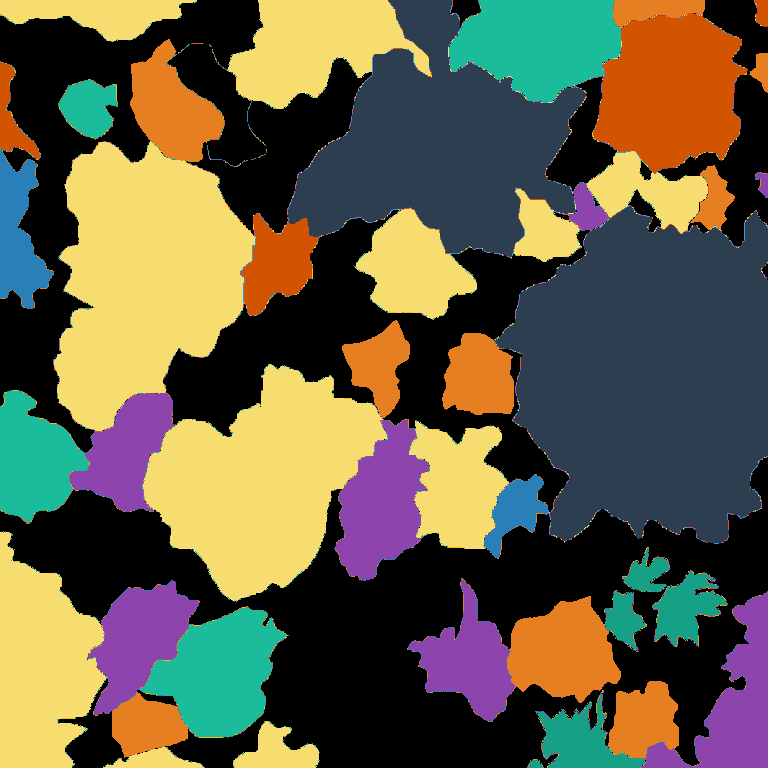}
        \caption{Annotation.}
        \label{fig:results_loss_2}
    \end{subfigure}
    \hfill
    \begin{subfigure}{0.22\textwidth}
        \includegraphics[width=\textwidth]{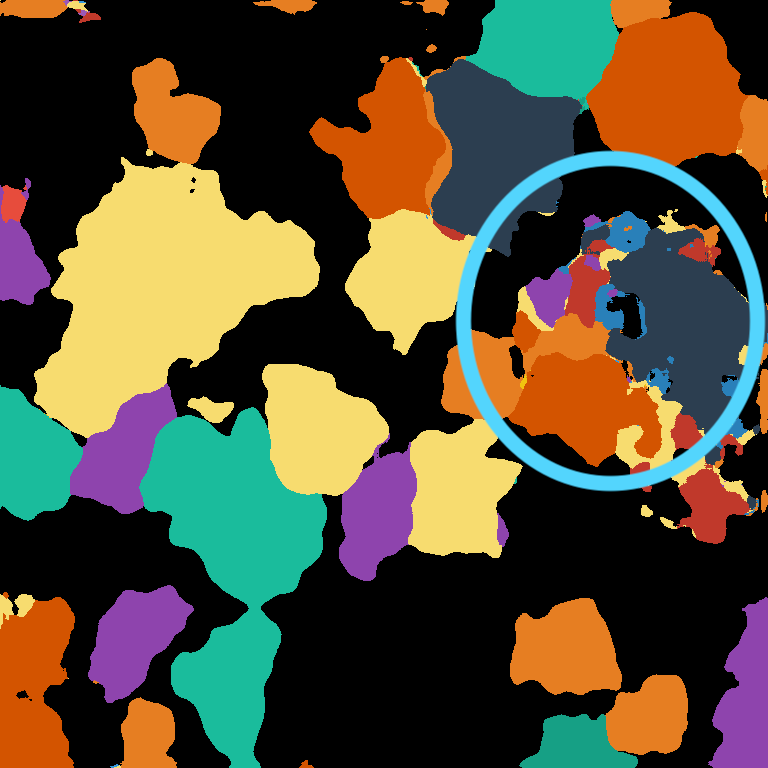}
        \caption{Results w/ Dice$+$CE.}
        \label{fig:results_loss_3}
    \end{subfigure}
    \hfill
    \begin{subfigure}{0.22\textwidth}
        \includegraphics[width=\textwidth]{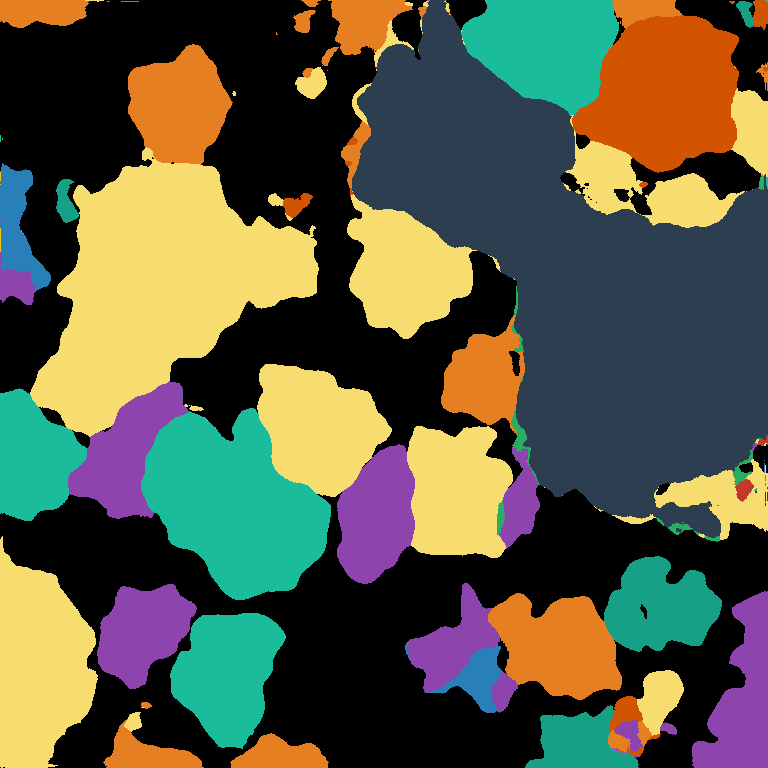}
        \caption{Results w/ HLoss.}
        \label{fig:results_loss_4}
    \end{subfigure}
    \caption{\textbf{Qualitative results of the Dice$+$CE loss versus HLoss.} This example compares the best-performing Processor$+$UNet (ResNet101) models trained with the Dice$+$CE loss and the proposed Hierarchical Loss (HLoss). First, \ref{fig:results_loss_1} shows a sample image from the sequence, while \ref{fig:results_loss_2} displays the corresponding ground truth annotation. Then, \ref{fig:results_loss_3} depicts the segmentation output obtained by the model trained with the Dice$+$CE loss, and finally \ref{fig:results_loss_4} illustrates the output from the model trained with HLoss. The colors of the labels and predicted segments correspond to specific tree species, as indicated by the legend in Table~\ref{tab:tree-abbreviations}. Upon closer inspection of the regions highlighted by the cyan circle (\protect\cyancirc{}), the model trained with the Dice$+$CE loss exhibits some confusion among classes, whereas the model trained with HLoss demonstrates improved discrimination between classes.}
    \label{fig:results_hloss_dice}
\end{figure}

\subsection{Single-image input for semantic segmentation}
\label{sec:results_single_image}

For the single-image segmentation model, we compare the performances of DeepLabv3+ and U-Net architectures with ResNet backbones of varying depths (ResNet-34, Resnet-50, ResNet-101) and the Mask2Former architecture with the Swin-T and Swin-S backbones. 

As seen in Table~\ref{tab:si_results}, both DeepLabv3+ and U-Net architectures show consistent increase in performances with increasing backbone size where the ResNet-101 model achieving the highest mIoU score. Moreover, the proposed HLoss consistently outperforms the Dice$+$CE loss across all backbones and architectures, demonstrating the effectiveness of leveraging taxonomic information. 
\begin{table}[tbp]
\centering
\begin{tabular}{lccc}
\toprule
\multicolumn{1}{c}{\textbf{Model}} & \multicolumn{1}{c}{\textbf{Backbone}} & \multicolumn{1}{c}{\textbf{Dice+CE}} & \multicolumn{1}{c}{\textbf{HLoss}} \\
\midrule

\multirow{4}{*}{DeepLabv3+ $+$ Processor} & ResNet34 & 53.32 $\pm$ 0.04 & \textbf{53.43 $\pm$ 0.34} \\
& ResNet50 & 53.12 $\pm$ 0.34 & \textbf{54.05 $\pm$ 0.31} \\
& ResNet101 & 53.46 $\pm$ 0.53 & \textbf{54.13 $\pm$ 0.11} \\
& ResNet50$^{\dagger}$ & 48.01 $\pm$ 0.37 & \textbf{50.30 $\pm$ 2.82} \\
\midrule
\multirow{4}{*}{U-Net $+$ Processor} & ResNet34 & 53.37 $\pm$ 0.53 & \textbf{53.60 $\pm$ 0.31} \\
& ResNet50 & 53.80 $\pm$ 0.15 & \textbf{54.12 $\pm$ 0.15} \\
& ResNet101 & 54.46 $\pm$ 0.39 & \textbf{\textcolor{red}{54.88 $\pm$ 0.20}} \\
& ResNet50$^{\dagger}$ & 49.00 $\pm$ 0.19 & \textbf{49.35 $\pm$ 0.25} \\
\midrule

\multirow{1}{*}{UNet 3D$^{\dagger}$} & --- & 37.74 $\pm$ 0.28 & \textbf{41.38 $\pm$ 0.14} \\
\midrule
\multirow{1}{*}{U-TAE$^{\dagger}$} & --- & 35.59 $\pm$ 1.03 & \textbf{39.78 $\pm$ 2.58} \\
\bottomrule
\end{tabular}
\caption{\textbf{Comparison of time-series methods with different losses and backbones.} Performances
are compared with IoU averaged over all the classes of the dataset (mIoU) for single image models. The $^{\dagger}$ indicates models trained from scratch. All the results are averaged over three seeds and the best results for a particular backbone is shown in bold text. The best model overall is highlighted in red.}
\label{tab:ts_results}
\end{table}

We also observe in Table~\ref{tab:si_results} that training models from scratch results in significantly lower mIoU scores compared to using pretrained ImageNet weights, highlighting the importance of transfer learning. The best performing single-image model is the U-Net with ResNet101 backbone. The Mask2Former models, trained with the loss of the original implementation and with pretrained weights from the MS-COCO dataset, perform better than the models trained from scratch, however their performance is not comparable to the CNN-based architectures.

\subsection{time series input for semantic segmentation}
\label{sec:results_time_series}

For time series inputs, we make use of the Processor module, detailed in Section~\ref{HLoss}, to extract spatio-temporal features and evaluate its performances with DeepLabv3+ and U-Net architectures. Amongst the time series models, we observe a similar pattern in Table~\ref{tab:ts_results} as the single-image ones with the benefical impact of using HLoss during training outperforming models trained with the Dice$+$CE loss. Qualitative results comparing HLoss with Dice$+$CE loss are illustrated in Figure~\ref{fig:results_hloss_dice} where HLoss demonstrates the ability to better discriminate between classes. Models trained using the Dice$+$CE loss exhibit some confusion among classes. Using HLoss would reduce confusion amongst classes that do not belong in the same genera or higher-level taxon as the model is penalized for incorrect predictions at all levels. The U-Net$+$Processor with ResNet-101 backbone trained with HLoss achieves the best mIoU score amongst all models. Furthermore, the time series models slightly outperform their single-image counterparts, indicating the importance of leveraging phenological patterns by incorporating temporal information for tree species segmentation. 

\begin{table}[tbp]
\centering
\begin{tabular}{lcc}
\toprule
\multicolumn{1}{c}{\textbf{Class}} & \multicolumn{1}{c}{\textbf{Processor + U-Net (R101)}} & \multicolumn{1}{c}{\textbf{U-Net (R-101)}} \\
\midrule
\multicolumn{3}{c}{\textbf{Non-Coniferous Trees}} \\
\midrule
\colorbox[HTML]{3498DB}{\textcolor{white}{Populus}} & \textbf{78.49} & 74.89 \\
\colorbox[HTML]{2980B9}{\textcolor{white}{ACPE}} & 29.64 & \textbf{29.74} \\
\colorbox[HTML]{8E44AD}{\textcolor{white}{ACRU}} & \textbf{57.51} & 55.84 \\
\colorbox[HTML]{27AE60}{\textcolor{white}{ACSA}} & \textbf{46.56} & 44.30 \\
\colorbox[HTML]{E74C3C}{\textcolor{white}{BEAL}} & \textbf{63.00} & 62.06 \\
\colorbox[HTML]{1ABC9C}{\textcolor{white}{BEPA}} & \textbf{72.55} & 71.16 \\
\colorbox[HTML]{F39C12}{\textcolor{white}{FAGR}} & 55.69 & \textbf{56.76} \\
% \colorbox[HTML]{16A085}{\textcolor{white}{DEAD}} & 0.4489 & 0.4823 \\
\midrule
\multicolumn{3}{c}{\textbf{Coniferous Trees}} \\
\midrule
\colorbox[HTML]{C0392B}{\textcolor{white}{PIST}} & 75.50 & \textbf{77.89} \\
\colorbox[HTML]{F7DC6F}{\textcolor{black}{Picea}} & 60.57 & \textbf{60.82} \\
\colorbox[HTML]{E67E22}{\textcolor{white}{ABBA}} & 62.35 & \textbf{64.15} \\
\colorbox[HTML]{D35400}{\textcolor{white}{THOC}} & \textbf{59.73} & 59.68 \\
\colorbox[HTML]{2C3E50}{\textcolor{white}{TSCA}} & \textbf{69.84} & 57.75 \\
\colorbox[HTML]{9B59B6}{\textcolor{white}{LALA}} & \textbf{76.68} & 76.60 \\
\midrule
\multicolumn{3}{c}{\textbf{Others}} \\
\midrule
\colorbox[HTML]{16A085}{\textcolor{white}{DEAD}} & 43.38 & \textbf{46.57} \\
\multicolumn{3}{c}{} \\
\midrule
\textbf{Overall results} & \textbf{54.88 $\pm$ 0.20} & 54.31 $\pm$ 0.48 \\
\bottomrule
\end{tabular}
\caption{The table shows the IoU for the individual classes for our best-performing Processor + U-Net and U-Net models, both with ResNet-101 as backbone. The classes are grouped into non-coniferous and coniferous categories, with the color shown for each class corresponding to the color code in Table \ref{tab:tree-abbreviations}. The last row presents the metrics from Table \ref{tab:si_results} and Table \ref{tab:ts_results} as a reference. These metrics represent the average performance across all classes over three seeds, not the average of the values shown in this table. We observe that incorporating time-series data improves the segmentation performance for most of the individual tree species. This performance gain is more pronounced for non-coniferous trees.}
\label{tab:class_wise_metrics}
\end{table}

To gain a deeper understanding of how leveraging time series data affects the performance of our models for individual species, we conduct a detailed analysis of the class-wise results for our best-performing single-image and time series models. For the single-image model, we select the U-Net architecture with a ResNet-101 backbone, while for the time series model, we choose the Processor+U-Net architecture, also with a ResNet-101 backbone. This allows for a fair comparison between the two approaches, as the main difference lies in the incorporation of temporal information through the Processor module. Table ~\ref{tab:class_wise_metrics} presents the class-wise Intersection over Union (IoU) scores for both models, with the classes grouped into non-coniferous and coniferous categories. Note that we omit a class from this analysis: 
% ``DEAD'', which represents dead trees and does not belong to either group, and 
``Acer sp.'', a class composed of trees belonging to ACPE, ACRU, or ACSA that have not been assigned a fine-grained label by the annotators due to low confidence.

The results show that the time series model consistently outperforms the single-image model across nearly all non-coniferous classes. Even in the few instances where the single-image model achieves a slightly higher IoU, the performance gap is minimal. This finding aligns with our hypothesis that incorporating time series data allows the models to better capture and exploit the phenological changes exhibited by different tree species, particularly those that undergo distinct color changes during the fall season. By leveraging this temporal information, the time series model is able to more accurately identify and distinguish between the various non-coniferous species.

\begin{figure}[tbp]
    \centering
    \begin{subfigure}{0.22\textwidth}
        \includegraphics[width=\textwidth]{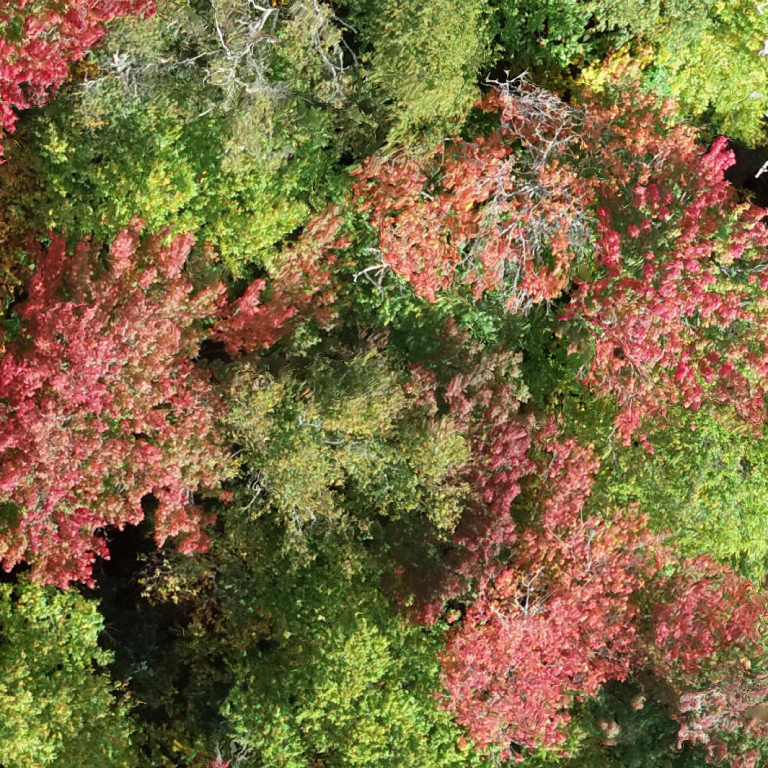}
        \caption{A sample image.}
        \label{fig:results_ts_1}
    \end{subfigure}
    \hfill
    \begin{subfigure}{0.22\textwidth}
        \includegraphics[width=\textwidth]{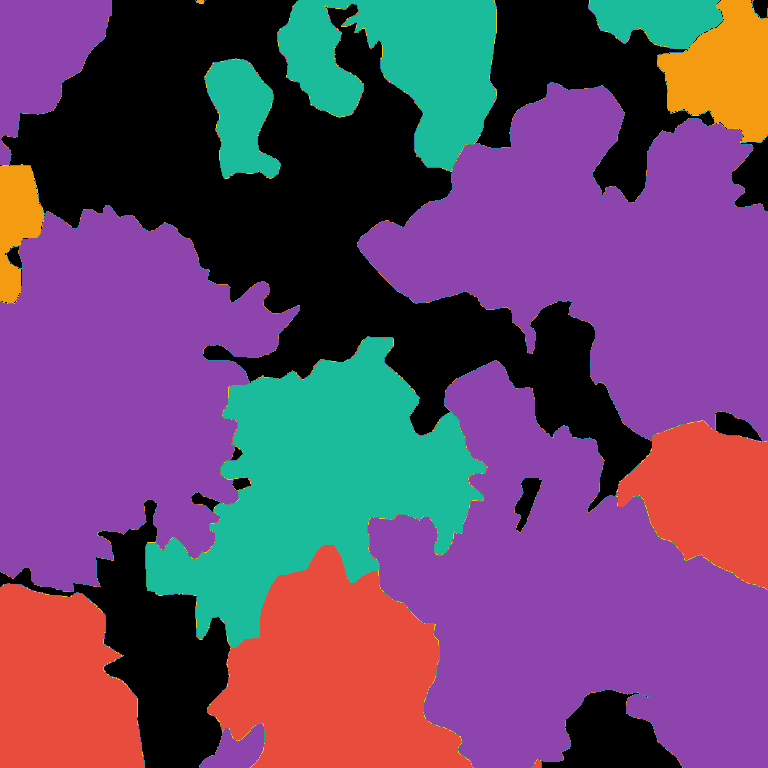}
        \caption{Annotation.}
        \label{fig:results_ts_2}
    \end{subfigure}
    \hfill
    \begin{subfigure}{0.22\textwidth}
        \includegraphics[width=\textwidth]{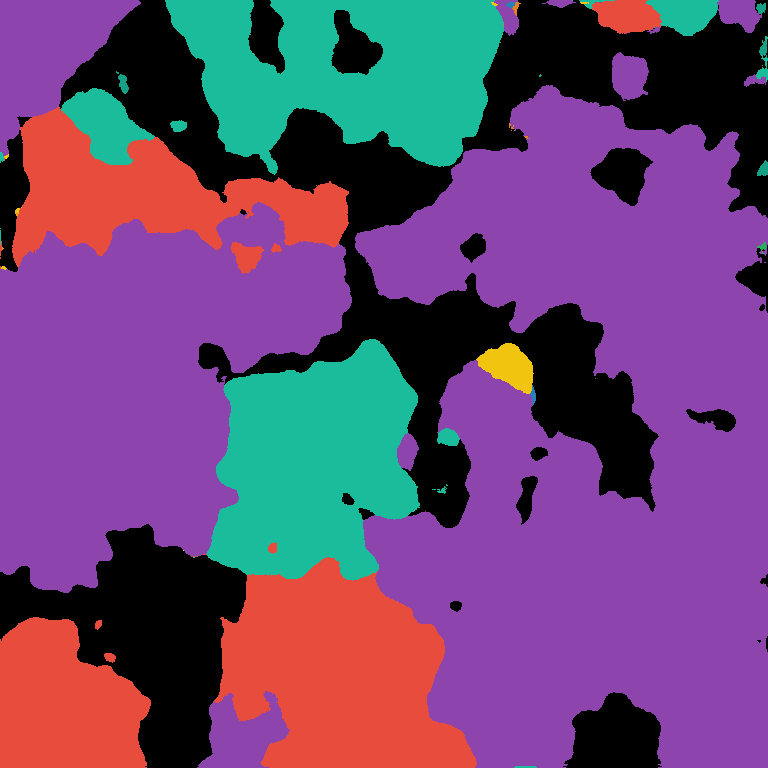}
        \caption{Results with SI}
        \label{fig:results_ts_3}
    \end{subfigure}
    \hfill
    \begin{subfigure}{0.22\textwidth}
        \includegraphics[width=\textwidth]{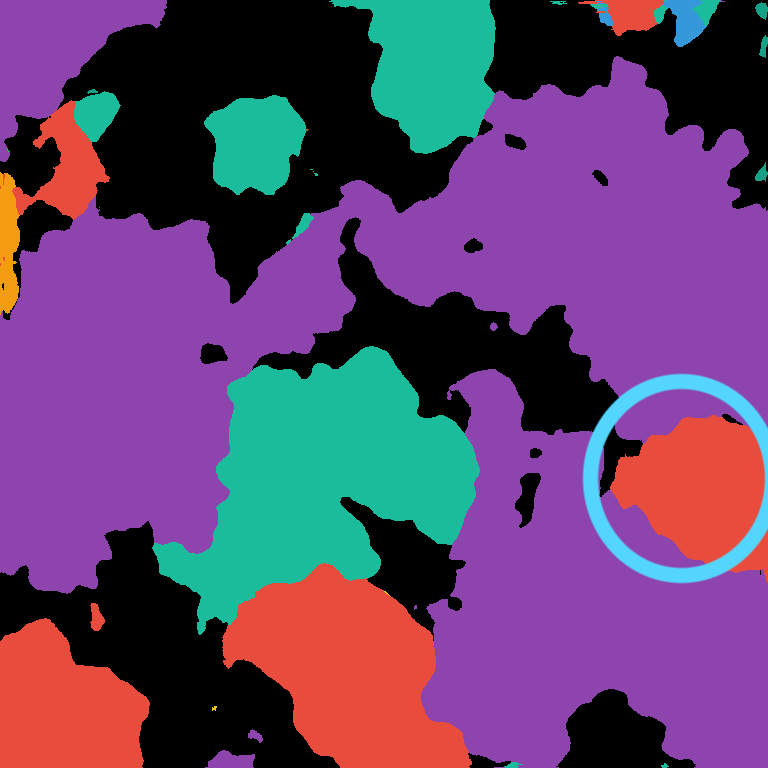}
        \caption{Results with TS}
        \label{fig:results_ts_4}
    \end{subfigure}
    \caption{\textbf{Qualitative results of the single-image versus time series inputs.} This example compares the best-performing models with single-image (SI) and time series (TS) inputs for tree species segmentation. The Processor$+$UNet (ResNet101) architecture, trained with the proposed Hierarchical Loss (HLoss), achieves the highest mIoU score according to Table~\ref{tab:ts_results}. First, \ref{fig:results_ts_1} shows a sample image from the sequence, while \ref{fig:results_ts_2} displays the corresponding ground truth annotation. Then \ref{fig:results_ts_3} depicts the segmentation output obtained by the single-image model, and finally \ref{fig:results_ts_4} illustrates the output from the time series model. The colors of the labels and predicted segments correspond to specific tree species, as indicated by the legend in Table~\ref{tab:tree-abbreviations}. Upon comparing the results, we observe that the time series model consistently outperforms the single-image model in correctly predicting the classes. In the instance highlighted by the cyan circle (\protect\cyancirc{}), the time series model accurately identifies the Swamp Birch, while the single-image model misclassifies it as Red Maple. }
    \label{fig:results_ts_main}
\end{figure}

In contrast, the performance differences between the single-image and time series models are less pronounced for the coniferous classes. Both models demonstrate strong performances in this category, with the most significant improvement for the time series model observed in the TSCA class. The single image model suffers from confusion between the classes Eastern hemlock (TSCA) and Eastern white pine (PIST) which does not affect the time series model as it takes in multiple input images with different lightning and acquisition angles. This suggests that even though phenological changes in coniferous trees may be less informative for species identification compared to their non-coniferous counterparts, the information from the image time series still helps the model identify the classes better.

An example of the results comparing single-image and time series models is illustrated in Figure~\ref{fig:results_ts_main}, where using temporal information helps the model differentiate between tree species that undergo senescence at slightly different times. Red maple trees are among the earliest trees to show color changes in the fall, and the single-image model misclassifies a Swamp Birch as Red Maple. This misclassification can be attributed to the lack of temporal context, which is necessary to understand the correlation between tree species and the timing of their senescence.

\section{Limitations}\label{limitations}

While our proposed Processor module demonstrates effectiveness in extracting spatio-temporal features for tree species segmentation, it has certain limitations that warrant further discussion. One notable drawback of the Processor is its inflexibility regarding the number of time steps in the input data. The current implementation assumes a fixed number of time steps, and any changes to the time series length would require retraining the module, although it is lightweight and easy to retrain by design. This lack of adaptability could pose challenges when exploring other datasets that have varying temporal resolutions. % or when incorporating additional time points becomes necessary.

Another potential area for improvement lies in the ability of the Processor to handle larger time series. As the number of time steps increases, the Processor might face difficulties to effectively capture and prioritize the most informative images within the sequence. Incorporating attention mechanisms into the Processor could help alleviate this issue by enabling the model to selectively focus on the most relevant time steps. By integrating attention into the Processor, the model could learn to assign higher weights to the time steps that contribute more significantly to the segmenting particular tree species, thereby improving its performance on larger time series datasets.

Despite these limitations, the Processor module offers a simple yet effective approach to leveraging temporal information in tree species segmentation. Its ability to be seamlessly integrated with existing segmentation architectures, such as U-Net and DeepLabv3+, makes it a versatile tool for enhancing the performance of these models. Moreover, the compact design of the Processor module allows for efficient computation and reduces the overall complexity of the model, making it suitable for resource-constrained scenarios.

\section{Conclusion}

In this work, we addressed the challenge of tree species segmentation using aerial image time series. We conducted a comprehensive evaluation of single-image and time series models, demonstrating the superiority of incorporating temporal information for accurately identifying tree species. Our results highlight the importance of leveraging phenological changes captured in time series data, as shown by the improved performances of time series models compared to their single-image counterparts.

We introduced a lightweight Processor module that effectively extracts spatio-temporal features from image time series, enabling the use of pretrained models for tree species segmentation. This module provides a flexible and efficient approach for incorporating temporal information into existing segmentation architectures. Additionally, we proposed a hierarchical loss function that leverages the taxonomic structure of tree species labels, allowing the models to learn more fine-grained distinctions while still benefiting from coarse-level information.

The proposed methods have significant implications for forest monitoring and biodiversity conservation, enabling accurate mapping of tree species composition, crucial for understanding forest ecosystems, monitoring changes over time, and informing conservation strategies. Future research could explore the incorporation of additional data modalities, addressing the limitations of the Processor module mentioned in Section \ref{limitations}, and the extension of the methods to other applications in forest ecology and management.

Our work demonstrates the potential of deep learning and time series analysis for advancing tree species segmentation and forest monitoring. By leveraging the rich information contained in aerial image time series and incorporating hierarchical knowledge, we can develop more accurate and efficient tools for understanding and preserving forest ecosystems in the face of global environmental challenges.

\begin{Backmatter}

\paragraph{Acknowledgments}
The authors are grateful for support from M. Cloutier for her guidance in understanding the intricacies of the dataset. In addition, we acknowledge material support from Mila Quebec AI Institute and from NVIDIA Corporation in the form of computational resources.

\paragraph{Funding Statement}
This work was funded through the IVADO program on "AI, Biodiversity and Climate Change" and the Canada CIFAR AI Chairs program.

\paragraph{Competing Interests}
None.

\paragraph{Data Availability Statement}
The dataset used in our work is published by the original authors here: 
\url{https://doi.org/10.5281/zenodo.8148479}. Our code can be found at \url{https://github.com/RolnickLab/Forest-Monitoring}.

\paragraph{Ethical Standards}
The research meets all ethical guidelines, including adherence to the legal requirements of the study country.

\paragraph{Author Contributions}

 Conceptualization: V.R.; A.O.; D.R. Methodology: V.R; A.O.; D.R. Data curation: V.R. Data visualisation: V.R. Writing original draft: V.R. Writing - Review \& Editing: V.R.; A.O.; D.R.. All authors approved the final submitted draft.

\bibliography{ref}

\end{Backmatter}

\end{document}